\documentclass[conference]{IEEEtran}
\IEEEoverridecommandlockouts
\usepackage{cite}
\usepackage{amsmath,amssymb}
\usepackage{graphicx}
\usepackage{textcomp}
\usepackage{authblk}
\usepackage[table,xcdraw]{xcolor}
\usepackage{booktabs}
\usepackage{hyperref}
\usepackage{algorithm}
\usepackage{algpseudocode}
\usepackage{verbatim} 
\usepackage{subfig}

\renewcommand{\baselinestretch}{0.95}
\usepackage{xcolor}
\usepackage{xcolor-material}
\usepackage{color, colortbl}

\def\BibTeX{{\rm B\kern-.05em{\sc i\kern-.025em b}\kern-.08em
    T\kern-.1667em\lower.7ex\hbox{E}\kern-.125emX}}
\begin{document}

\title{Fault-Aware Design and Training to Enhance DNNs Reliability with Zero-Overhead

}

\author[1]{Niccol\`o Cavagnero} 
\author[2]{Fernando dos Santos}
\author[1]{Marco Ciccone}
\author[1]{Giuseppe Averta}
\author[1]{Tatiana Tommasi}
\author[3]{Paolo Rech}
\affil[1]{Department of Control and Computer Engineering, Polytechnic of Torino, Italy}
\affil[2]{Univ Rennes, INRIA, France}
\affil[3]{University of Trento, Italy}


\maketitle
\begin{abstract}
\label{abstract}
Deep Neural Networks (DNNs) enable a wide series of technological advancements, ranging from clinical imaging, to predictive industrial maintenance and autonomous driving. However, recent findings indicate that transient hardware faults may corrupt the models prediction dramatically. For instance, the radiation-induced misprediction probability can be so high to impede a safe deployment of DNNs models at scale, urging the need for efficient and effective hardening solutions. In this work, we propose to tackle the reliability issue both at training and model design time. First, we show that vanilla models are highly affected by transient faults, that can induce a performances drop up to 37\%. Hence, we provide three zero-overhead solutions, based on DNN re-design and re-train, that can improve DNNs reliability to transient faults up to one order of magnitude. We complement our work with extensive ablation studies to quantify the gain in performances of each hardening component.
 
\end{abstract}


\section{Introduction}
\label{sec:introduction}

Deep Learning is more and more pervasive in our daily lives, with the number of AI-based applications sharply increasing and the deployment of intelligent systems becoming ubiquitous. We count a number of novel technologies that are enabled by machine learning, ranging from diagnosis of malignancies, to automatic predictive maintenance of industrial machines, to fully autonomous vehicles. While the advantages of this trend are tautological, the potential harm due to the adoption of this technology should not be underestimated. As an example, it has been observed that machine learning methods are highly prone to adversarial attacks which, in some instances, may completely change the desired behaviour of neural networks~\cite{adversarial}. Additionally, machine learning methods have been shown to be prone to radiation-induced soft errors~\cite{santosTR2019, Ibrahim2020}. The probability of experiencing a radiation-induced corruption during interference is exacerbated by the large size and complexity of the hardware required to execute Deep Neural Network (DNN) models. 
Despite the low error rate per device (in the order of one error every 3-4 years, considering a natural flux of 13~$neutrons/cm^2/h$~\cite{Jedec2006}, for modern GPUs~\cite{Ibrahim2020, tc2016}), the foreseen large-scale adoption of DNNs in vehicles (10s of millions cars on the move in the EU, on the average) and Internet of Things applications (billions of devices connected), make DNNs reliability evaluation and improvement mandatory.
\begin{figure}[t]
    \centering
    \includegraphics[width=0.86\columnwidth,keepaspectratio]{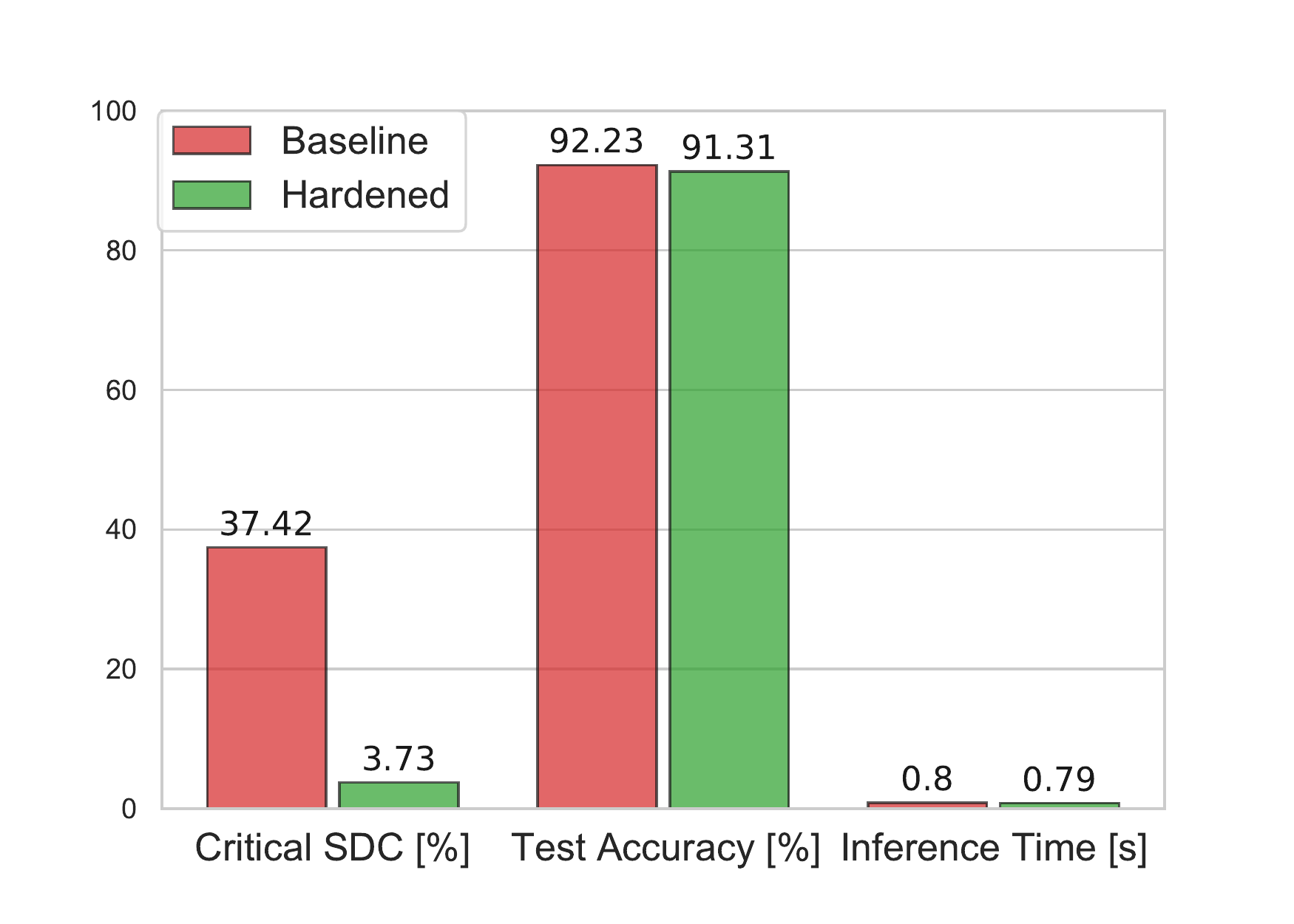}
    \caption{Critical SDCs [\%], Test Accuracy [\%] in fault-free execution, and Inference Time (for batch size 200)~[s] for Baseline and Hardened CIFAR10. The Hardened model shows a significantly higher reliability to transient faults while roughly preserving the accuracy with no execution time overhead.}
    \vspace{-0.5cm}
    \label{fig:init}
\end{figure}
In this scenario, hardware protection is too costly and most of the available software solutions to improve reliability are either very inefficient (replication) or simply adapt to DNNs strategies derived from the protection of classical algorithms (e.g. selective hardening~\cite{Mahmoud2021, Libano2019} or check-sums in convolutions~\cite{santosTR2019, Mittal2020}). In this paper, we propose to change the reliability paradigm for DNNs, exploiting their potentialities and using intrinsic features of DNNs designs. Our goal is to reduce as much as possible the impact of transient faults without affecting the network performances.

Recent advancement of machine learning research demonstrated that training and design of neural models can effectively increase the robustness of DNNs to a variety of noise typologies or adversarial attacks~\cite{adversarial}, without impacting performances. Inspired by these findings, we investigate whether -and to which extent- a proper re-design of existing deep architectures, complemented with a fault-aware training, can recover the accuracy drop caused by soft errors induced by ionising particles strikes. 

Our results, teased in Figure~\ref{fig:init}, demonstrate that a combination of three actions at training time, with zero overhead at test time and almost no accuracy loss, is able to considerably reduce the mispredictions induced by soft errors (from more than 37.5\% to 3.7\% in the case of CIFAR10). We provide an incremental ablation study to clarify the role of each contribution we propose. 
Ultimately this paper advances the State of the Art with the following: 
\begin{itemize}
    \item we discuss and quantify the effects of soft errors in image classification tasks performed with standard DNNs;
    \item we propose three independent solutions that can increase the robustness of neural models to hardware faults;
    \item we investigate and validate the contribution given by each proposed solution and provide a proper combination to fully exploit their potential. With high-level (Python) and low-level Source And Assembly (SASS) fault injections we show that our solutions reduce of up to one order of magnitude the probability of mispredictions with zero inference time overhead.
\end{itemize}
\section{Background}
\label{sec:background}

In this section, we provide a brief background on DNNs, specifically focused on the concepts we exploit to train and design more reliable networks. For each possible design choice we also highlight its impact in the DNN reliability. Then, we discuss the potential effects of radiations on computing devices and DNNs.

\subsection{Deep Neural Networks Design and Training}

Neural Networks (NN) are universal function approximators~\cite{approx} that, thanks to Backpropagation training~\cite{backpropagation} and a sufficient network complexity, enable the solution of a variety of tasks, e.g. classification, detection, and regression. 
Interestingly, DNNs, and Convolutional Neural Networks (CNNs) in particular, efficiently map in parallel processors and, therefore, benefit from the heavy usage of GPUs computation for both the training and inference processes.

The \textbf{design} process of DNNs consists of the identification of the number and typology of layers that, once properly interconnected, can be adapted to the specific task. The  adaptation is performed through a \textbf{training} phase, in which the parameters of the network are modified in such a way that, for each training input-output pairs, the network response is as close as possible to the ground-truth. The distance measured between true and predicted values is called loss function.

The design choices, together with the methods that are used to train the network, strongly impact the overall performances of the model in solving the desired task. The network design is responsible for the expressivity and the trainability of the architecture, i.e. its capability to encode the knowledge required by the task. The training oversees an effective tune of all the network parameters. Only a judicious combination of proper techniques can result in a neural architecture capable of solving the task with good performances. Additionally, as we show in this paper, only a proper design/training can make the DNN intrinsically more reliable to transient faults.

Each network design has a specific set and organisation of layers. 
\textbf{Convolutional layers} have different hyperparameters, specifically kernel size, stride and padding. Each kernel is independent and produces a different feature map, with as many output feature maps as the number of filters. 

Besides convolutions (that are layers that are most computationally demanding and, thus, vulnerable to radiation), \textbf{activation functions} are used for ensuring a non-linear input-output relationship in DNNs and are very often implemented through Rectified Linear Units (ReLU)~\cite{relu} 
%
   $ ReLU(x) = max(0, x),$
where $x$ stands for the input tensor.
This definition for activation layers enables an easy gradient flow, which is fundamental for the \textbf{Backpropagation} operation performed during training~\cite{deep}. Building upon this function, several other ReLU-like activations have been developed, e.g. SELU~\cite{selu}, GELU~\cite{gelu}, ReLU6~\cite{mobilenetv2}. 
\textbf{Normalisation layers} play a role in the stabilisation of neural architectures training, by smoothing the optimisation landscape~\cite{batchnorm} and by preventing the weight and gradient explosion.
The most common normalisation layer is \textit{BatchNorm}, which learns at training time an approximation of the first and second statistic moments of each feature map
to normalise the input tensors. After the normalisation, it is standard practice to apply an \textit{Affine transform}, namely an additive bias $\beta$ and a scale parameter $\gamma$. The normalisation operation can then be defined as:
\begin{equation}
\setlength\abovedisplayskip{2pt}
BatchNorm(x) = \frac{x - \mathbb{E}[x]}{\sqrt{\mathrm{Var}[x] + \epsilon}} * \gamma + \beta,
\setlength{\belowdisplayskip}{3pt}
\end{equation}
where $\mathbb{E}[x]$ is the expected value of the input tensor $x$, $\mathrm{Var}[x]$ its variance and $\epsilon$ is a correction to improve stability~\cite{deep}.
In vanilla CNNs, the usual layer order is convolution-normalisation-activation, but other design options have been proposed in the last years. For example, in \cite{preresnet} the activation function is moved before the convolution (pre-activation), while in \cite{vit} the first operation is the normalisation (pre-norm). However, there is no consensus in the community and different architectures benefit from different choices in the layer ordering. Of note, in this paper we show that this also plays a role in the robustness of the model to transient faults.

The process of tuning the network parameters to fit the dataset is called training and is usually performed via \textbf{Backpropagation} of the output error from the last layer all the way back to the input one. First, a batch of data is forwarded through the network and the output is compared with the ground-truth labels by means of a loss function, e.g. Cross Entropy or L2. Since neural networks implement differentiable operations only, it is possible to compute the gradients of the loss w.r.t. the network weights. Given the gradients, an optimiser, e.g. Stochastic Gradient Descent~\cite{sgd}, update the weights in order to minimise the loss function. This forward and backward steps are repeated for each batch of training data for a certain number of epochs (i.e. a complete pass over the whole training set). It is of the utmost importance to use as training set a highly heterogeneous set of data, because this enhances the generalisation capabilities of the model. Indeed, DNNs generally suffer a significant performance drop when deployed to scenarios they were not trained for. For this reason, it is standard practice to use heavy Data Augmentations strategies to obtain a more varied training set that can contain useful information not present in the original data, e.g. change light conditions if the original training set presents day scenes only. We apply a similar approach to transient faults.

\subsection{Radiation effects on computing devices}

It is well known that radiation can induce \textit{transient} faults in the hardware that can (1) be \emph{masked} without affecting the software, (2) lead to a \textbf{Silent Data Corruption} (SDC) (incorrect application output) or (3) generate a \textbf{Detected Unrecoverable Error (DUE)}, which is a crash or a device reboot. In this paper we will focus just on SDCs, that are particularly problematic, as their silent nature makes them extremely hard to be detected and can potentially lead to unstable or unknown system states. 
Previous studies have already investigated the reliability of DNNs executed on parallel and programmable devices through radiation experiments~\cite{santosTR2019, Libano2019, rubens2022TNS, sulivanMicro2021NVIDIA} and fault injections~\cite{hariCNNHARD2021, li2017SC, nvbitfi2021, Ruospo2020}.

Lately, various hardening solutions have been proposed with the aim of reducing the impact of transient faults in DNNs. Algorithm-Based Fault Tolerance (ABFT)~\cite{abftInria2018,santosTR2019, hariCNNHARD2021}, filters to detect propagating errors in MaxPool layers~\cite{santosTR2019, li2017SC} and (selective) replication with comparison~\cite{rpdwc2022, Libano2019} have been shown to significantly increase DNNs reliability. While these solutions have been shown to be effective, they are sub-optimal since they do not exploit DNN programming philosophy but rather adapt hardening solutions, derived from classical computation, to DNNs. 
Zahid \emph{et. al}~\cite{fatUssama2020} propose a first attempt of a fault-aware training for Quantised Neural Networks (QNNs) and Hoang \emph{et al.}~\cite{Hoang2020} propose to clip ReLU values to improve reliability. However, the proposed techniques are limited to QNNs in FPGAs or are limited to single bit-flips and are not suitable for DNNs and complex accelerators like GPUs. 
With this paper we intend to move a step forward in the quest of efficient reliability of DNNs by taking advantage of network training and network design.

Independently on the DNN model and on the underlying hardware architecture, previous works have shown that: (i) not all SDCs are critical for a DNN, since some output errors still allow the correct detection/classification; (ii) the convolution tends to spread the hardware fault: a single bit flip can corrupt a significant portion of the feature map; (iii) the value of the transient corruption (i.e., how much the corrupted output is different from the correct output) is neither random nor can be simplified with a single bit flip model.

To have an effective DNN hardening it is paramount to select a realistic fault model to present at training time. Using a synthetic fault model (e.g. single bit flip) would lead to unrealistic evaluations and hardening solutions that result ineffective, once employed in the field. In this paper, as a case study, we consider the recently published fault model, based on beam experiments and RTL injections, for convolutions executed in GPUs~\cite{Lunardi2018, santosTR2019, santosDSN2021}. What has been highlighted is that a single transient fault spreads and corrupts multiple elements of the convolution output. These corrupted elements are distributed in a block, on a line, or randomly distributed. 

\subsection{Main Idea and Contribution}

Our idea is to make a DNN more reliable proposing improvements at the design/train stage. We intend to \textit{choose the DNN design} (activation functions, layers order) that is more likely to reduce the number of misclassifications caused by hardware faults but still guarantees the highest performances. Additionally, we perform \textit{fault-aware training} to improve the DNN ability to properly classify the objects even in the event of a transient fault. By injecting faults in specific moments of the training process we force the DNN to learn how to properly deal with the most critical errors.

\section{DNNs Reliability Improvement Methodologies}
\label{sec:methodology}
We propose three non intrusive methods, derived from the knowledge on DNN models and on previous experimental observations, to improve the reliability of DNNs, with zero overhead at inference time.
\subsection{Activation Function}
\label{sec:activation}
It has been showed that most of the critical faults for DNNs (i.e., the faults that cause misclassification or misdetection) typically modify significantly the values propagated through the network~\cite{santosTR2019, rangeRestrict}. Intuitively, given the intrinsic approximation of DNNs, if the corrupted value is close to the correct one we do not expect major output corruptions. Previous works have introduced specific layers in the DNN with the sole role to detect these high-magnitude faults~\cite{rangeRestrict}. While this solution is effective, it requires to add extra layers, reducing the network performances in terms of computations and inference speed. We show that this is not necessary, since DNNs already have intrinsic features to filter excessive values. 

\textbf{The first reliability improvement we propose is to replace the standard ReLU activation with its clipped counterpart, ReLU6~\cite{mobilenetv2}, and train the DNN} (dissimilarly to~\cite{Hoang2020}) with this new activation function. ReLU6 is a ReLU in which all values greater than six are clipped to six:
\begin{equation}
    ReLU6(x) = min(max(0, x), 6).
\end{equation}
This activation was originally proposed for mobile devices, with the aim of improving the quantisation of the weights. Indeed, ReLU6 was successfully applied in various Computer Vision tasks such as classification and detection~\cite{mobilenetv2}.

It is worth noting that using ReLU6 is \textit{not} the same as reducing the precision of the DNN operations. ReLU6 does not force the network to use only values from 0 to 6, but simply clips the values that reach the layer. In other words, convolutions can still output values outside of [0; 6] and the network can still have weights outside of [0; 6] but, when the values propagate to the ReLU6 layer they are clipped. By training the DNN with ReLU6 we ensure to force the DNN to reduce as much as possible the use of values outside the bounds while not loosing performances.

Our intuition is that ReLU6 also naturally allows DNNs to significantly reduce the impact of transient faults. Indeed, even if a neutron strike could in theory perturb a feature map generating high-magnitude values, these would be simply scaled down to six, reducing the impact of the error and possibly allowing the DNN to recover from this perturbation with much less effort. Our experimental results shown in Section~\ref{sec:results} confirm this intuition.

\subsection{Fault-Aware Training}
\begin{figure*}[t]
    \centering
    \includegraphics[width=0.8\linewidth]{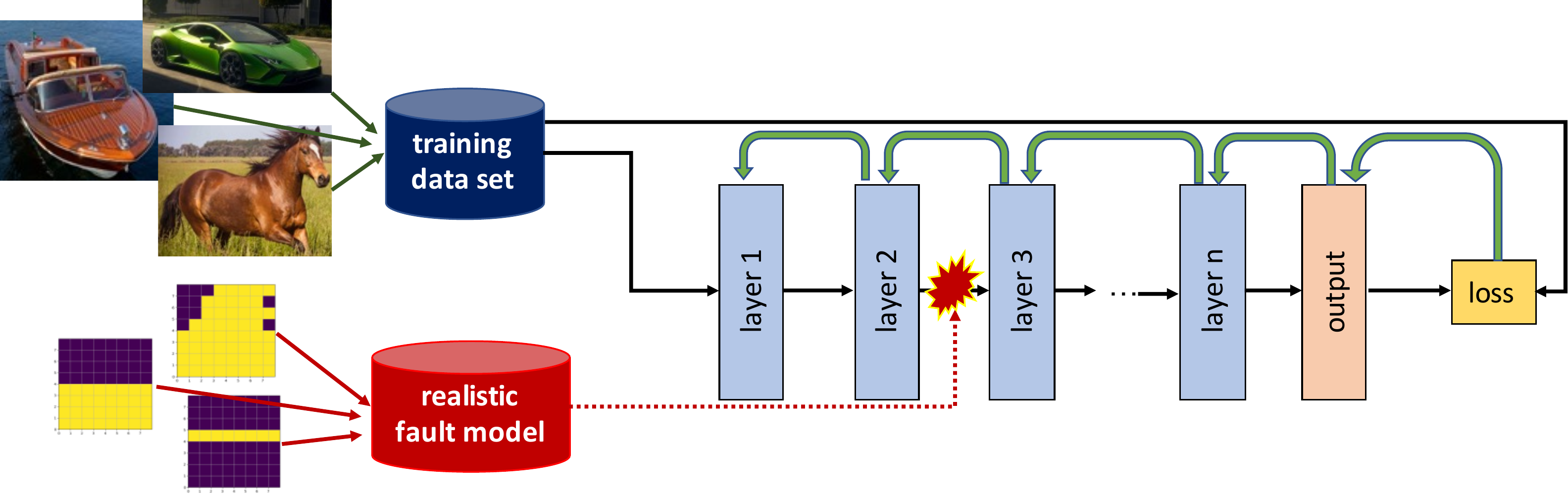}
    \caption{Scheme showing the fault-aware training pipeline. We randomly select the fault model to apply at the output of a random convolution layer. During a training, we perform $\approx$ 3.75 millions of injections.}
    \vspace{-0.5cm}
    \label{fig:fi}
\end{figure*}
DNNs are very powerful tools to perform the tasks they have been trained for. However, DNNs usually perform poorly when deployed in scenarios not seen during the training phase. For instance, a model trained exclusively with day scenes is going to experience a significant drop in performances when tested on night scenes with poor illumination. The most common solution to address this problem is to use a dedicated Data Augmentation (DA) strategy (i.e. solarisation or greyscaling in this case) that enables the network to also experience situations not present in the original training data. We propose to adopt the very same strategy to transient faults. Our intuition is that, if the DNN is trained to properly classify objects even with some selected transient faults, we could produce a more reliable DNN, maintaining the original performances.

\textbf{Our second reliability improvement is to allow the DNN to familiarise with the occurrence of neutron-induced errors by injecting noise (transient faults) during training} (see Section~\ref{train_details} for details about injections). In particular, as shown in Fig.~\ref{fig:fi}, while performing the forward pass of the DNN training, we randomly corrupt the feature maps (convolution output) in a given layer of the network, injecting a realistic fault model. As a result, we allow the model itself to autonomously learn how to properly deal with this kind of fault by adjusting the learned weights in order to reduce the likelihood of a misprediction.
The challenge to address is the selection of the faults to present to the DNN in the training phase. Considering a high number of faults could increase the network experience in dealing with faults, but risks to prevent the training convergence. A low number of faults will result in a quick training but might be ineffective. Moreover, we cannot be sure that a certain random fault in a given layer is going to lead to an error (thus, the DNN will learn how to deal with it). From our experience, injecting single bit flips, for instance, is transparent to the network training. The approach we adopt is to identify the fault models that are more likely to induce mispredictions and inject these faults in the majority of the training samples (see Section~\ref{train_details} for more details). During training, injections lead to a limited time overhead of $\approx$ 20 minutes per training, increasing the training time from $\approx$ 50 minutes to $\approx$ 70.

\subsection{DNN Design}
\textbf{The third contribution we propose is to re-design the DNN to minimise the probability of fault propagation.} We have observed that in vanilla neural networks each computational block is typically composed by an ordered juxtaposition of convolution, normalisation and activation operations. In the case of noise affecting the inner layers of neural architectures, as for the soft errors discussed in this paper, the probability of fault occurrence in the convolution operation is higher than the other operators, since the first is by far the most computing-expensive operation~\cite{yolov3}. When a corrupted sample is forwarded within the network, at the feature level this appears as an outlier w.r.t. the distribution of features. As a consequence, training with these outliers can alter the learned statistics in the normalisation layers, with a negative impact on performances even for clean data. Therefore, to limit this possibility, it is reasonable to expect that a clipping of features through ReLU6 \textit{before} the actual computation of normalised features can not only limit the error propagation but also reduce the role of single outliers in the distribution (i.e. corrupted samples) during training.
\begin{figure}[b!]
    \centering
    \vspace{-0.5cm}
    \includegraphics[width=0.54\columnwidth]{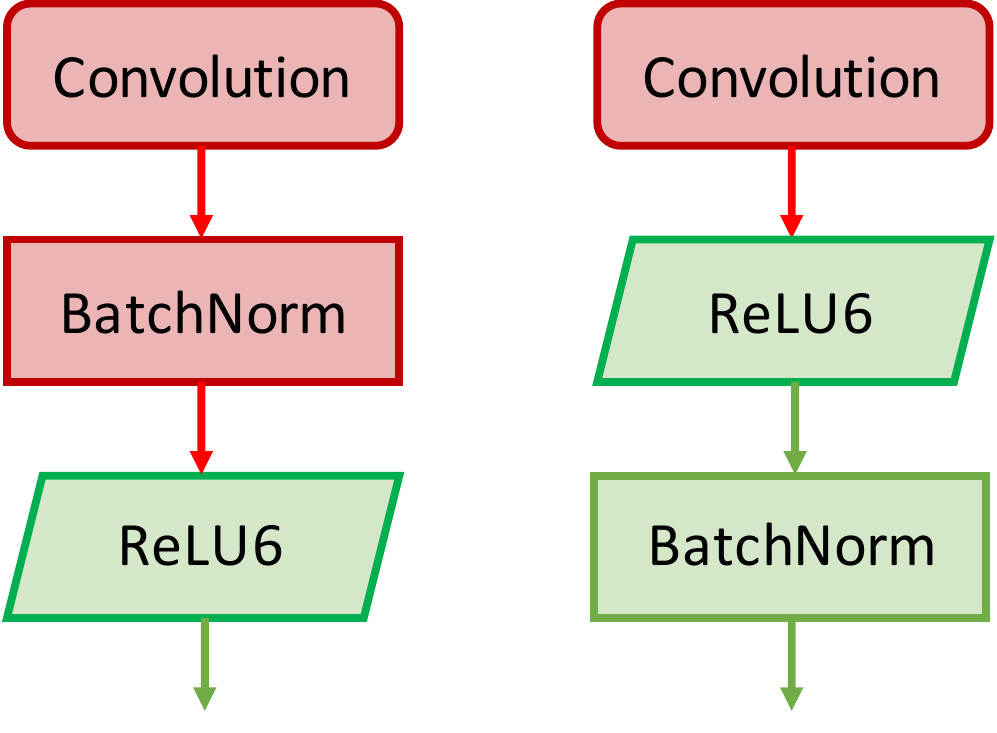}
    \caption{Scheme of the operation order for a vanilla DNN (left) and our proposed order (left). Red: layers with high probability of fault propagation. Green: layers with cropped errors.}
    \label{fig:order}
\end{figure}
As discussed in Section~\ref{sec:activation}, ReLU6 can greatly reduce the impact of high-magnitude errors. By inverting the order of normalisation and activation layers, as shown in Figure~\ref{fig:order}, ReLU6 is directly applied after the most critical and compute-hungry operation (i.e., convolution). The same layer ordering was adopted, for example, in~\cite{mixer}. Thus, the error propagation is hindered before feeding the feature maps to normalisation layers, which are then able to learn more accurate statistics and a proper Affine transform. Our results demonstrate that, without this architectural change, the improvement given by the fault-aware training may be limited by poorly learned statistics (see Section~\ref{sec:results}).

\section{Hardening and Evaluation Methodology}

\subsection{Case Study}
While our methodology can be applied to any network, hardware architecture, and fault model, to evaluate the effectiveness of the proposed DNNs hardening solution we have selected, as a case study, a standard ResNet44~\cite{resnet} trained from scratch on two common Computer Vision benchmarks, namely CIFAR10 and CIFAR100. The two datasets contain 50,000 32x32 images for training and 10,000 for testing with 10 different classes in the former and 100 in the latter. We choose Titan V GPUs as supporting hardware, and choose the open-source transient fault model available in~\cite{dsn2021REPO}.

To show the impact of the three proposed actions, we performed an incremental ablation study: first, we report the results of a standard ResNet44 with ReLU and with ReLU6, then we perform a fault-aware training with the ReLU6 architecture and finally we invert the order of normalisation and activation. To evaluate the reliability of the hardening solution we report Test Accuracy for the fault-free and faulty (``noisy'') execution. The drop in performance (accuracy) between the two is hereinafter named regret. We further validate the improved reliability with a low-level fault injection.

\subsection{Fault Injections}
\label{fi}

To inject faults during the DNN training, we use a specially crafted Python-level fault injector, which is extremely fast (nearly zero-overhead). Since the training process is very time consuming, injecting errors using low-level fault injectors is unfeasible. During training we inject faults at the output of convolution. To ensure effective hardening solutions, we inject, the convolution fault models that we have observed with beam experiments or RTL fault injections~\cite{Lunardi2018, santosTR2019, santosDSN2021}. In other words, we tracked the transient fault propagation from the hardware to the manifestation at the output of a convolution and inject these during training. The fault model consists of the geometry observed on the output tensor of a convolution (e.g., if the fault corrupted a single line or multiple lines of the tensor) and the statistical distribution of the incorrect values (how much the values diverge from the expected value).
The fault model used for this work is available at~\cite{dsn2021REPO}. It is worth noting that, as faults are injected in software, if we inject a single bit flip (i.e., we assume that the hardware fault results in a single bit corrupted in the output of a convolution), we would train the DNN to deal with a na\"{i}ve and unrealistic fault model. The resulting DNN would then be hardened against a much less critical fault than the realistic one. 

After the fault-aware training, we validate the proposed DNN reliability improvements solutions through both the high-level fault injection via Python and with a Source And Assembly level fault injection using NVBitFI~\cite{nvbitfi2021}. 
SASS-level injections are more realistic than application-level injections as we can simulate errors in the microinstructions of the DNN. However, the SASS-level injections on large applications such as DNNs impose a high overhead (i.e., in the order of minutes for a single injection). This is impractical to be integrated in the complex training process (we inject thousands to millions faults for a single training set). 
We inject two fault models on SASS-level fault injection, Single Bit Flip on the floating point instruction output and Warp Random Values. On Warp Random Value injection, we select all the threads within a GPU warp (i.e., the smallest thread execution group on a GPU) and replace the output of a float instruction with a random value following a power-law distribution~\cite{santosDSN2021}.

\subsection{Training Details}
\label{train_details}
Since our goal is to mitigate the neutron-induced faults effect, rather than achieving the maximum possible Test Accuracy, we trained for 100 epochs without applying other data augmentation strategies beyond random crop and horizontal flipping. We adopted a standard Stochastic Gradient Descent~\cite{sgd} optimiser, a Cosine Annealing~\cite{cosine} scheduler and a Binary Cross Entropy loss. The initial learning rate is set to 2, we used a batch size of 128, weight decay of 1e-5 and gradient clipping of 1.
When performing fault-aware training, for each batch of images which is forwarded in the network we first sample the convolution or linear layer which is affected by the injection. Once the batch is forwarded to the chosen layer, we sample 0.75 of the images which the noise will be applied to. At this point, we sample the error magnitude (sampled from a Uniform [0, epoch], i.e. we linearly increase the maximum possible magnitude during training) and the channels which are affected by the injection. At training time only ``block'' errors, i.e. errors which affect a random block in the feature map, are injected, while when testing we randomly choose between single value, line and block errors. In case of single value noises all channels are corrupted, while in case of line noises the probability of channel perturbation is 0.75 and 0.3 for the block ones. The same procedure is adopted when validating with high-level injections, with the difference that at test time all images in a batch are corrupted.
\begin{figure}[t]
    \centering
    \includegraphics[width=0.9\columnwidth,keepaspectratio]{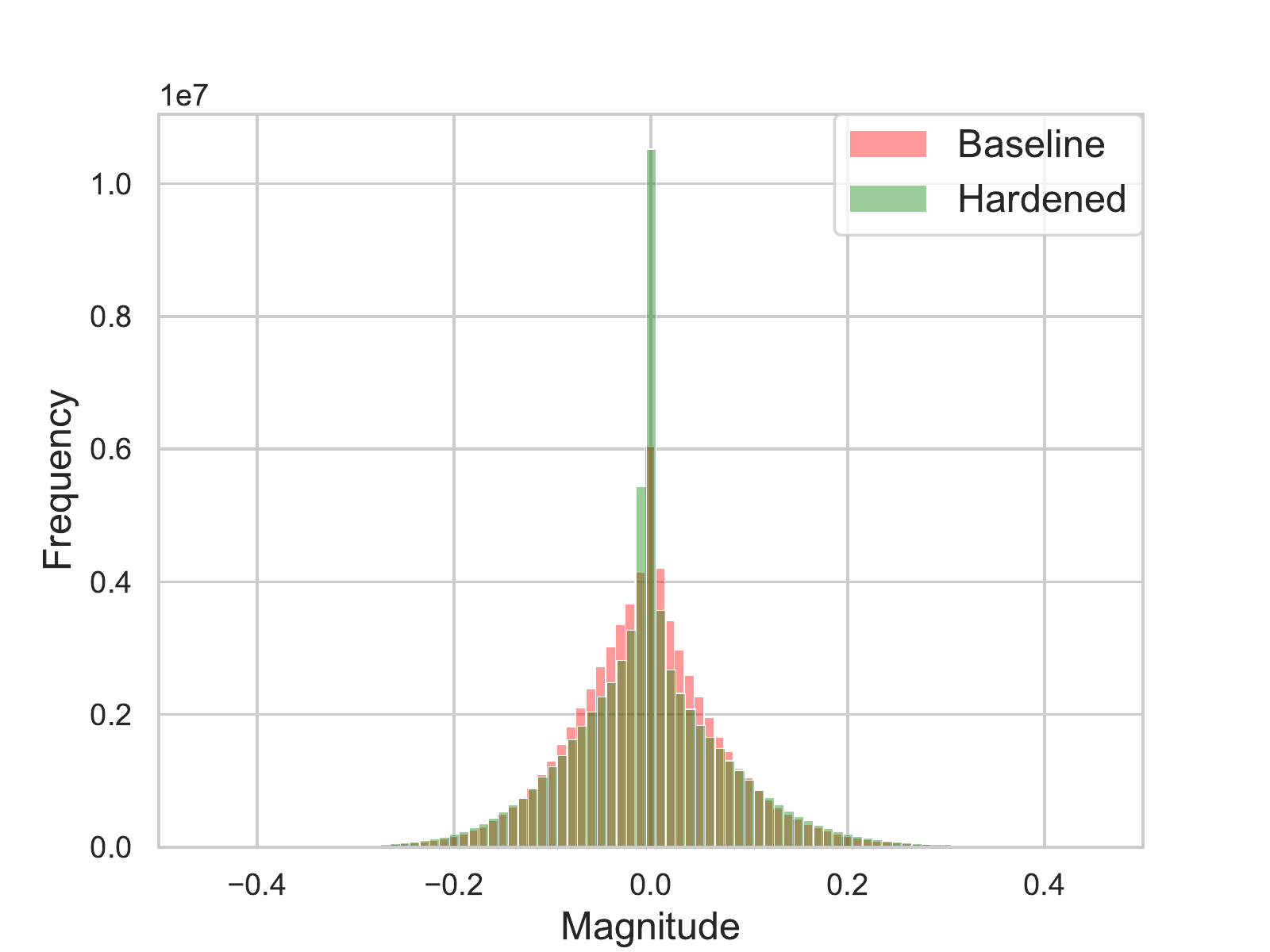}
    \caption{Weights distribution for Baseline and Hardened models on CIFAR10. X-axis is restricted in [-0.5, 0.5] for visualisation purposes. A similar plot can be drawn for CIFAR100.}
    \label{fig:dist}
    \vspace{-0.5cm}
\end{figure}

\section{Results}
\label{sec:results}
As teased in Fig.~\ref{fig:init} on the CIFAR10 dataset our \textit{hardened} model shows no overhead at inference time and a with a very limited accuracy drop in the fault-free case. Similarly, the hardened CIFAR100 experience a negligible drop in accuracy in the fault-free case, from 69.53\% to 69.48\%.
Fig.~\ref{fig:dist} shows the weight distribution of the \textit{hardened} model shows visible differences w.r.t. the baseline. On average, the hardened weights have half the magnitude (-4.13e$^{-3}$ vs -2.31e$^{-3}$ on CIFAR10, -1.72e$^{-3}$ vs -9.78e$^{-4}$ on CIFAR100) and are much more concentrated around zero). Hence, the fault-aware training modified the DNN weights and the \textit{hardened} model is less affected from perturbations in the feature maps, whose propagation is limited not only by ReLU6 but also by the lower magnitude of the weights.

\subsection{High-level Fault Injections}
\begin{figure}[t]
    \centering
    \includegraphics[width=0.88\columnwidth,keepaspectratio]{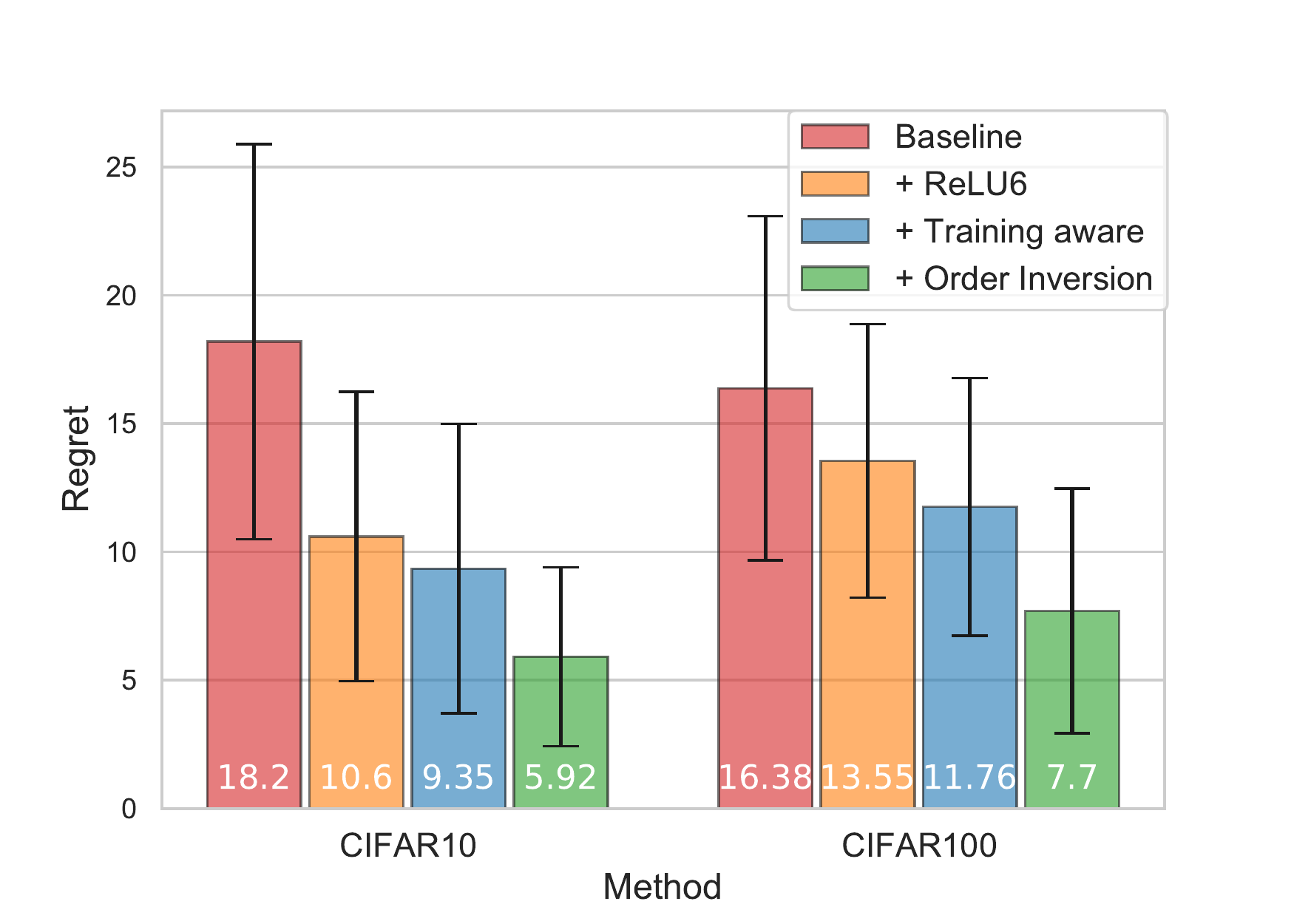}
    \caption{Average Regret per configuration across 5 different injections per sample on CIFAR10 (left) and CIFAR100 (right).}
    \label{fig:ablation}
    \vspace{-0.5cm}
\end{figure}
As a first evaluation to understand the behaviour of the different configurations we consider high-level injections. Specifically, we measure the Regret, i.e. the difference in Test Accuracy for the fault-free and the faulty (``noisy'') cases. The baseline ResNet on average shows an accuracy drop of more than 15\% on both the considered dataset. By replacing the ReLU activation with ReLU6, we provide an already significant protection with a $\approx 8\%$ and $3\%$ improvement on CIFAR10 and CIFAR100 respectively. This result supports our hypothesis that ReLU6 can limit the magnitude and propagation of the error in the architecture and can therefore play a crucial role in the design of robust DNNs. Interestingly, the improvement given by the fault-aware training alone is very limited (see Fig.~\ref{fig:ablation}). We claim that this is a consequence of poorly learned statistics in the normalisation layers due to the perturbations induced by noise-injections at training time. This behaviour is strongly related to the architecture structure which prescribes normalisation before activation. In other words, in this vanilla design the normalisation acts on non-clipped features, thus leading to improper learned statistics and Affine transform. Indeed, by changing the order of normalisation and activation layers, we obtain an increase of Noisy Test Accuracy up to $\approx$ 12\% and 9\% on CIFAR10 and CIFAR100 respectively. This attests that fault-aware training alone might not be sufficient, and the network design must also be considered to maximise the training effect.

\subsection{SASS-level Fault Injections}

\begin{figure}[t]
    \centering
    \includegraphics[width=1\columnwidth,keepaspectratio]{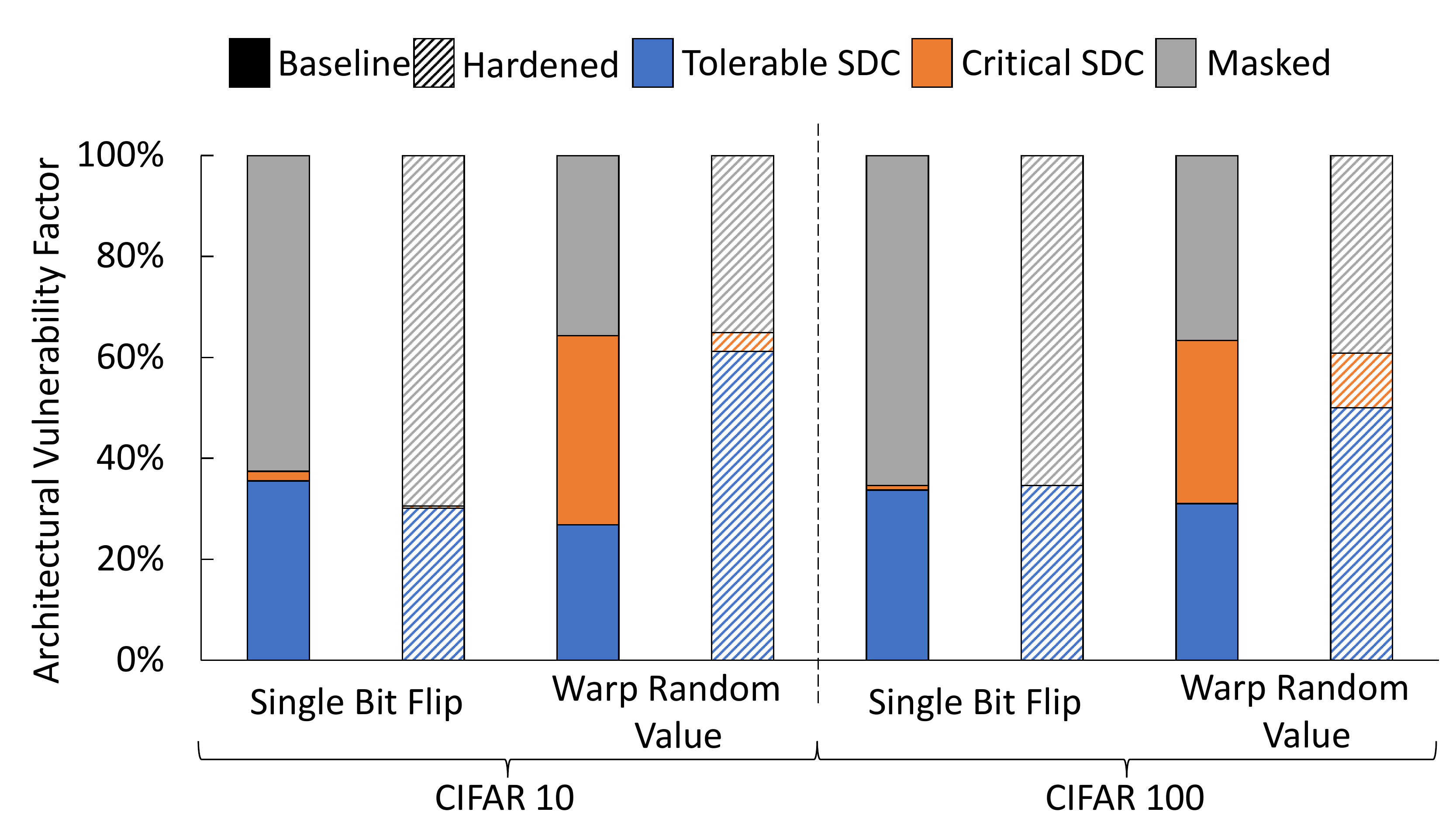}
    \caption{AVF for CIFAR10 and CIFAR100 in the Baseline and Hardened (ReLU6+Training aware+Order inversion) versions.}
    \label{fig:cifar_avf}
\end{figure}

Figure~\ref{fig:cifar_avf} depicts the Architectural Vulnerability Factor (AVF) for the SASS-level fault injections performed with NVBitFI. We select two low-level fault models to validate the hardened DNNs, the Single Bit Flip and Warp Random Value. Both fault models have been observed in low-level instructions on GPUs~\cite{santosDSN2021}. The main difference between the Python and SASS injections is that the former inject the faults after convolution is completed, the latter allows the fault to propagate from the machine instruction till the output of convolution and of the DNN.

To have a detailed evaluation we divide the fault outcomes in three categories, \emph{Tolerable SDCs}, \emph{Critical SDCs}, i.e., mispredictions, and fault that are \emph{Masked}. We limit the low-level fault injection to the Baseline and the most advanced Hardened network (ReLU6 + Training aware + order inversion). Figure~\ref{fig:cifar_avf} shows that the hardened DNN has a much lower probability to experience Critical SDCs (mispredictions) for both datasets, CIFAR10 and CIFAR100.

For Single Bit Flip injections the hardened DNNs has almost zero mispredictions while for the Warp Random Value injections the AVF for Critical SDCs is reduced of more than 1 order of magnitude for CIFAR10 and of $3.2\times{}$ on CIFAR100. This result, obtained with a low-level fault-injection, further validates the achieved reliability of the hardened DNNs.

Interestingly, the overall number of SDCs (Tolerable+Critical) is basically the same between the Baseline and Hardened DNN. This suggests that the injections modify the DNN execution and, in both the Baseline and Hardened DNN, the fault is manifested at the output. However, in the Hardened DNN the network is able to deal with the fault and still produce the correct prediction. This observation attests that the improved reliability is not achieved thanks to a higher fault masking but rather to a better DNN design.
\section{Conclusions}
\label{sec:conclusions}

In this paper we have exploited DNNs potentiality and the knowledge about the fault generation and propagation to reduce the number of radiation-induced mispredictions. The three proposed hardening solutions we propose include the use of a clipped activation function, the fault-aware training, and the re-design of the DNN. These solutions are very effective in improving the DNN ability to deal with transient faults since, as shown with both high-level and low-level injections, we can reduce of 1 order of magnitude the occurrences of mispredictions. The most interesting contribution relies on the lack of overhead imposed by the hardening strategies, since the interference time remains unaltered. We strongly believe that our findings opens to the possibility of adapting to reliability various solutions of DNN design improvements. In the future we plan to study more advanced DNN architectures, investigating data augmentation strategies.

\section*{Acknowledgments}

This project has partially received funding from the European Union’s Horizon 2020 research and innovation program under the Marie Sklodowska-Curie grant agreement No 886202 and 899546 and with the support of the Brittany Region.

\bibliographystyle{IEEEtran}
\bibliography{IEEEabrv,references}

\end{document}